\documentclass[conference]{IEEEtran}
\IEEEoverridecommandlockouts
\usepackage{cite}
\usepackage{amsmath,amssymb,amsfonts}
\usepackage{algorithmic}
\usepackage{graphicx}
\usepackage{textcomp}
\usepackage{booktabs}
\usepackage{array} 
\usepackage{cuted}
\usepackage{multirow} 
\usepackage[flushleft]{threeparttable} 
\usepackage{afterpage}
\usepackage{color, soul}

\usepackage{xcolor}

\def\BibTeX{{\rm B\kern-.05em{\sc i\kern-.025em b}\kern-.08em
    T\kern-.1667em\lower.7ex\hbox{E}\kern-.125emX}}

\begin{document}
%
\title{Suitability of KANs  for Computer Vision:\\ A preliminary investigation}

\author{
\IEEEauthorblockN{Basim Azam, Naveed Akhtar}
\IEEEauthorblockA{School of Computing and Information Systems\\
The University of Melbourne, Australia\\
Email: \{basim.azam, naveed.akhtar1\}@unimelb.edu.au}
}

\maketitle

\begin{abstract}
Kolmogorov-Arnold Networks (KANs) introduce a paradigm of neural modeling that implements learnable functions on the edges of the networks, diverging from the traditional node-centric activations in neural networks. This work assesses  the applicability and efficacy of KANs in visual modeling, focusing on fundamental recognition and segmentation tasks. We mainly analyze the performance and efficiency of different network architectures built using KAN concepts along with  conventional building blocks of convolutional and linear layers, enabling a comparative analysis with the conventional models. Our findings are aimed at contributing to understanding the potential of KANs in computer vision, highlighting both their strengths and areas for further research. Our evaluation\footnote{This manuscript is a work in progress, which includes results from an ongoing exploration in this direction. The results and discussion are expected to evolve over time. This independent effort is intended to contributes to our understanding of KANs for visual modeling, and not to provide a yes/no answer to the question `if KANs should replace conventional neural modeling techniques for computer vision'.} 
point toward the fact that while KAN-based architectures perform in line with the original claims, it may often be  important to employ more complex functions on the network edges to retain the performance advantage of KANs on more complex visual data.
\end{abstract}


%
\IEEEpeerreviewmaketitle

\section{Introduction}

Neural networks have been foundational in advancing the field of machine learning, particularly in the tasks requiring complex pattern recognition such as image and speech processing \cite{lecun2015deep, haykin2009neural,schmidhuber2015deep}. Traditionally, Multi-Layer Perceptrons (MLPs) have been a cornerstone in neural network architectures \cite{geidarov2017clearly,diep2023crossmixed}. MLP is a fully connected network consisting of multiple layers of nodes, each applying a nonlinear activation function to a weighted sum of inputs from the previous layer\cite{hornik1989multilayer, cybenko1989approximation}. This configuration has proved potent in approximating nonlinear functions across various domains \cite{leshno1993multilayer,pinkus1999approximation}.

MLPs are often challenged by the demands of high-dimensional data, such as images \cite{lecun1998gradient}. This stems from their structure, which lacks the capacity to inherently capture spatial patterns in the data, resulting in scalability and efficiency concerns. Consequently researchers investigate  architectures that can handle such complexities more effectively. Convolutional Neural Networks (CNNs)~\cite{lecun1989backpropagation} have emerged as a powerful class of neural networks that are particularly suited for analyzing visual imagery \cite{krizhevsky2012imagenet,simonyan2014very}. Building upon the foundational principles of MLPs, CNNs adapt and extend basic concepts to more effectively model the intricate structures of input data. CNNs are designed to automatically and adaptively learn spatial hierarchies of features through backpropagation \cite{he2016deep, szegedy2015going}. This capability makes them  effective for tasks such as image recognition\cite{krizhevsky2012imagenet}, object detection \cite{ren2015faster,he2017mask}, and segmentation \cite{long2015fully,ronneberger2015u}.

The recent proposition of Kolmogorov-Arnold Networks (KANs)~\cite{liu2024kan,vaca2024kan, de2024kan} introduces an innovative modification to the traditional neural networks by employing  learnable functions on the edges of the graph rather than using fixed activation functions at the nodes. This architecture is inspired by the Kolmogorov-Arnold Representation Theorem \cite{schmidt2021kolmogorov}, which posits that any multivariate continuous function can be decomposed into univariate functions and a summing operation. In KANs, each element within the layer applies a unique, learnable function to its inputs. In their seminal work, Liu et al.~\cite{liu2024kan} claimed that KANs can address some of the intrinsic limitations of MLPs, particularly in handling complex functional mappings in high-dimensional spaces.

Building on theoretical foundations of KAN, the application of this type of neural architectures to visual data presents a novel area of study. Given the structured nature of image data and the critical role of efficient function approximation in image recognition and classification, KANs could offer significant advantages. Preliminary attempts of ConvKANs \cite{AntonioTepsich2024} (the convolutional adaptation of KANs) have shown promising results by replacing the traditional dot product in convolution operations with a learnable non-linear activation, adapting the flexible and powerful Kolmogorov-Arnold theorem directly into the processing of visual information.

Several implementations and research efforts are currently being made to explore the potential of KANs in various domains. 
The Convolutional-KANs project \cite{AntonioTepsich2024} extends the idea of KANs to convolutional layers, replacing the classic linear transformation of the convolution with learnable non-linear activations in each pixel. The Vision-KAN repository~\cite{chenziwenhaoshuai2024} explores the possibility of MLPs with KANs in Vision Transformers. The KAN-GPT project \cite{AdityaNG2024} implements Generative Pre-trained Transformers (GPTs) using KANs for language modeling tasks, demonstrating the potential of KANs in natural language processing applications.
The KAN-GPT-2 repository \cite{CG804992024} trains small GPT-2 style models using KANs instead of MLPs. The KANeRF project \cite{Tavish92024} integrates KANs into Neural Radiance Fields (NeRF) for view synthesis tasks. 
The KAN-UNet repository~\cite{JaouadT2024} implements a U-Net architecture with Kolmogorov-Arnold Convolutions  for image segmentation tasks. Similarly, the kansformers project~\cite{akaashdash2024} explores the use of KANs in Transformer architectures, replacing the traditional linear layers with KAN layers.

Although there is currently a large   interest in developing implementation repositories providing various KAN variants  pertinent to visual modeling, a formal scientific effort focusing on the suitability of KANs in general and the emerging repositories in particular for visual modeling tasks is still missing. 
This work is aimed at partially addressing this gap as a preliminary effort focusing on providing independent empirical evidence of KANs' suitability for fundamental vision task. 
We replicate KAN-based visual modeling ideas 
to evaluate their performance in terms of accuracy, training efficiency, and model parameters while experimenting with standard datasets. 
Our experiments are expected to provide insights into the strengths and limitations of KANs in handling complex visual data. The study aims at helping understand answers to the following questions:

\begin{enumerate}
    \item How do KANs perform in terms of accuracy, training efficiency, and model parameters in the context of visual modeling?
    \item Can KANs be effectively integrated into existing CNN frameworks to enhance their performance or efficiency?
    \item What are the potential limitations or challenges in adapting KANs to complex visual tasks?
\end{enumerate}

This manuscript is structured to provide an assessment of KANs' abilities for visual modeling (considering image recognition task). Following this introduction, we present a theoretical formulation of KANs and their extension to ConvKANs. We then describe the experimental setup, implementation details, and results of applying KANs to standard datasets. The subsequent sections provide  discussion on the validation of claims regarding KANs in \cite{liu2024kan}, followed by a critical analysis of their suitability for the broader applications in computer vision. The manuscript concludes with a summary of the findings and directions for future research. 
{This document will be periodically updated as new data becomes available and further experiments are conducted.}

\section{Theoretical Formulation}
\begin{figure*}[t]
    \centering
    \includegraphics[width = 0.7\textwidth]{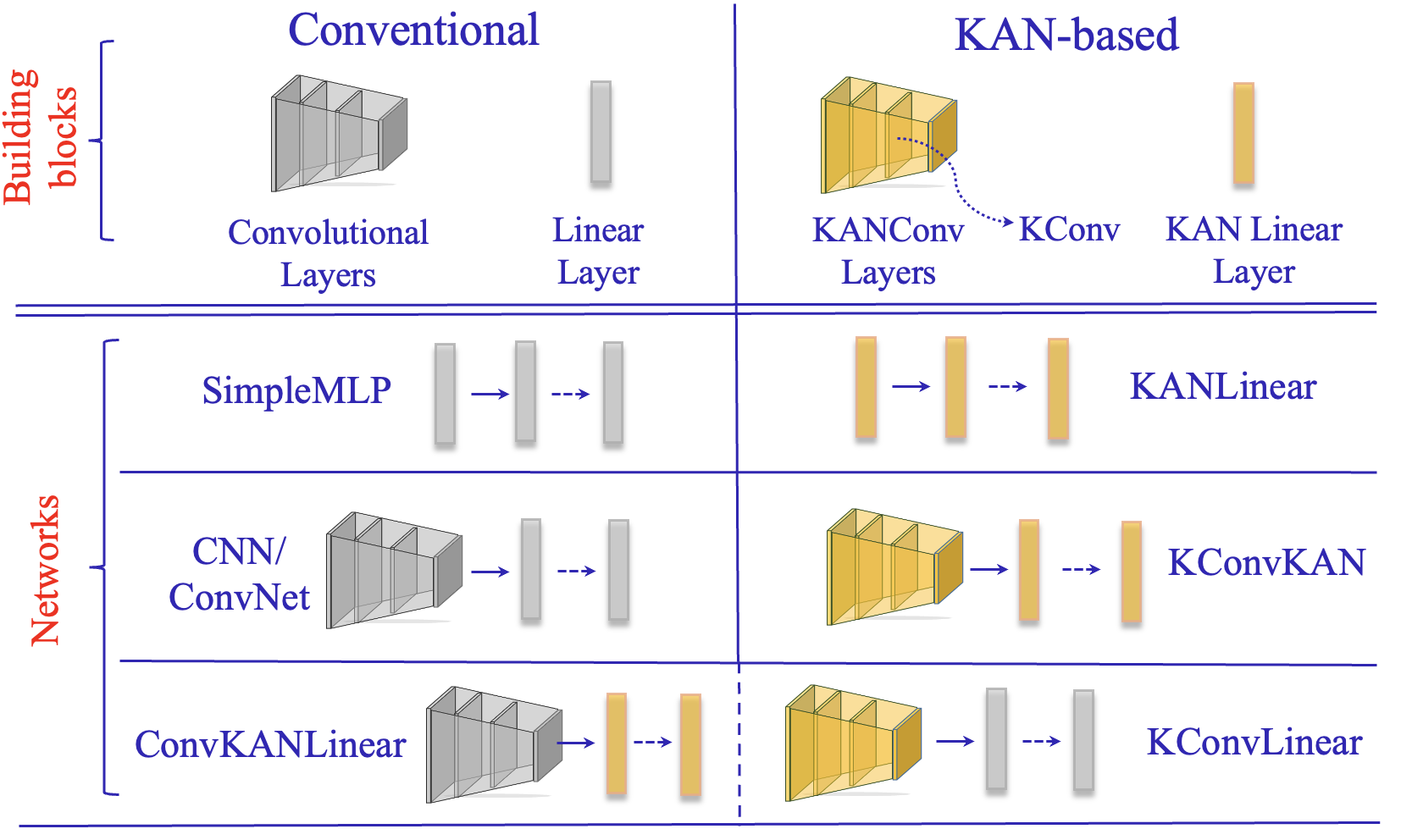}
    \caption{Categorization of the types of network architectures used in this work. We employ KAN-based building blocks with conventional layers to construct different types of networks. The same naming conventions are used throughout this work.}
    \label{fig:Nom}
\end{figure*}

This section explores the mathematical foundations and theoretical constructs of KANs~\cite{liu2024kan} and their adaptation into convolutional neural networks \cite{AntonioTepsich2024}.
The foundational novelty of KANs is in learning parametric functions on graph edges. This concept can be used in traditional fully-connected layers of a network as well as convolutional layers. We use ConvKAN as a general term for the convolutional networks leveraging  KAN concepts.
When both convolutional and linear layers are  implemented with KANs, we often refer to the network as KConvKAN to particularly emphasize on the usage of KAN-based convolutional layers. Figure~\ref{fig:Nom} provides the nomenclature used  in this manuscript for a quick reference.  


\subsection{Multi-Layer Perceptrons (MLPs)}

A Multi-Layer Perceptron (MLP) is a fully connected feedforward neural network consisting of multiple layers of nodes (neurons). Each node in a layer is connected to every node in the subsequent layer, and the node applies a nonlinear activation function to the weighted sum of its inputs.

Universal Approximation Theorem \cite{hornik1989multilayer} provides foundations to the popularity of MLPs. The theorem  states that a feedforward network with a single hidden layer containing a finite number of neurons can approximate any continuous function on compact subsets of $\mathbb{R}^n$, given appropriate activation functions.
Mathematically, an MLP with $L$ layers can be expressed as:
\begin{equation}
\begin{aligned}
\operatorname{MLP}(\mathbf{x}) = \sigma_L(\mathbf{W}_L \sigma_{L-1}(\mathbf{W}_{L-1} \cdots \sigma_1(\mathbf{W}_1 \mathbf{x} + \mathbf{b}_1) \cdots \\
+ \mathbf{b}_{L-1}) + \mathbf{b}_L),
\end{aligned}
\end{equation}
where:
\begin{itemize}
    \item $\mathbf{x}$ is the input vector.
    \item $\mathbf{W}_l$ and $\mathbf{b}_l$ are the weight matrix and bias vector for the $l$-th layer.
    \item $\sigma_l$ is the activation function for the $l$-th layer.
\end{itemize}

\subsection{Kolmogorov-Arnold Networks (KANs)}
Here, we present the concepts related to KANs by building on the work of Liu et al.~\cite{liu2024kan}. Interested readers are referred to  \cite{liu2024kan} for more details. 
\subsubsection{Kolmogorov-Arnold Representation Theorem}
The Kolmogorov-Arnold Representation Theorem \cite{kolmogorov1957representation} states that any multivariate continuous function can be represented as a finite composition of continuous functions of a single variable and the binary operation of addition. Specifically, for a smooth function $f: [0,1]^n \rightarrow \mathbb{R}$,
\begin{equation}
f(\mathbf{x}) = \sum_{q=1}^{2n+1} \Phi_q \left( \sum_{p=1}^{n} \phi_{q,p}(x_p) \right),
\end{equation}
{where $\phi_{q,p}$ and $\Phi_q$ are continuous functions and $x_p$ denotes the $p^{\text{th}}$ component of the input vector $\textbf{x}$.} 
\subsubsection{KAN Architecture}
KANs generalize the Kolmogorov-Arnold representation by using learnable activation functions on graph edges. A KAN layer with $n_{\text{in}}$-dimensional inputs and $n_{\text{out}}$-dimensional outputs is defined as:
\begin{equation}
\mathbf{\Phi} = \left\{ \phi_{q,p} \right\}, \quad p=1,2,\ldots,n_{\text{in}}, \quad q=1,2,\ldots,n_{\text{out}}
\end{equation}
where $\phi_{q,p}$ are learnable functions, parameterized as splines.
The output of a KAN layer is given by:
\begin{equation}
\mathbf{x}_{l+1} = \sum_{i=1}^{n_l} \phi_{l,j,i}(x_{l,i}), \quad j=1,\ldots,n_{l+1}.
\end{equation}
In matrix form it can be expressed as:
\begin{equation}
\mathbf{x}_{l+1} = \boldsymbol{\Phi}_l \mathbf{x}_l,
\end{equation}
where $\boldsymbol{\Phi}_l$ is the function matrix corresponding to the $l^{\text{th}}$ KAN layer.
KANs are claimed to possess faster neural scaling laws than MLPs \cite{liu2024kan}. In particular, the approximation bound for KANs is given by:
\begin{equation}
\left\| f - \left( \boldsymbol{\Phi}_{L-1}^G \circ \cdots \circ \boldsymbol{\Phi}_0^G \right) \mathbf{x} \right\|_{C^m} \leq C G^{-k-1+m},
\end{equation}
where $G$ is the grid size, $k$ is the order of the B-spline, and $C$ is a constant. We followed the notational convention from \cite{liu2024kan}, where $\boldsymbol{\Phi}_{L-1}^G \circ \cdots \circ \boldsymbol{\Phi}_0^G$ denotes the composition of the functions matrices from layer 0 to layer $L-1$. The composition operation $\circ$ indicates the output of one funciton matrix is used as the input to the next function. 

\subsection{Extension to ConvKANs}

\subsubsection{Convolutional Neural Networks (CNNs)}
Convolutional neural networks~\cite{lecun1989backpropagation} are a class of deep neural networks that use convolutional layers to process data with a grid-like topology, such as images. This makes them ideally suitable for computer vision tasks. A convolutional layer applies a set of learnable filters (kernels) to the input, producing feature maps.
Mathematically, the output feature $y_{i,j}$ computed at location $(i,j)$ on the featured grid by a convolutional kernel $k$ is expressed as:
\begin{equation}
y_{i,j} = \sum_{m,n} x_{i+m,j+n} k_{m,n}.
\end{equation}

\begin{figure*}[hbtp]
    \centering
    \includegraphics[width=0.8\textwidth]{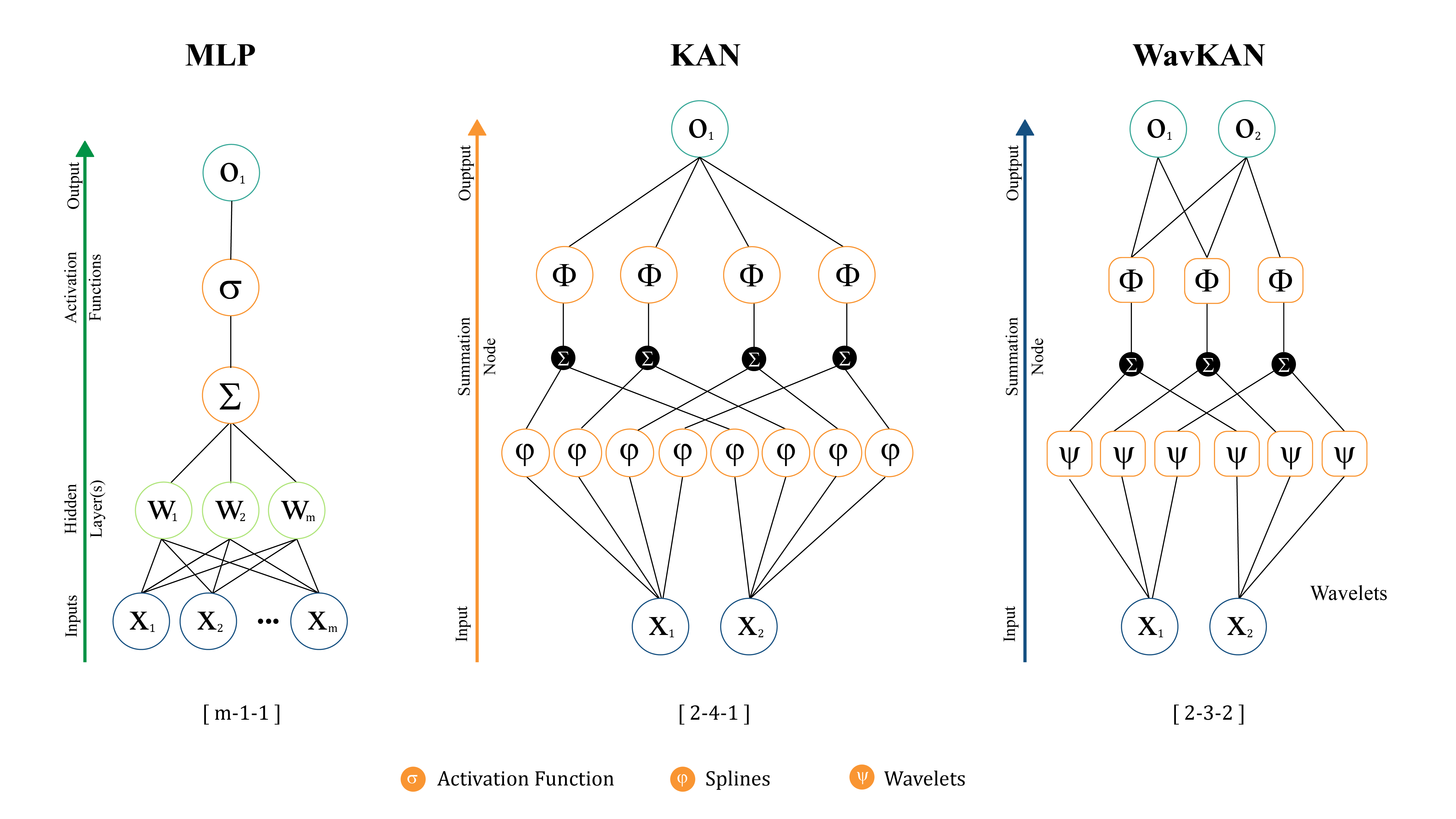}
    \caption{\textcolor{black}{A high-level comparison of basic network configurations using Multi-Layer Perceptrons (MLP), Kolmogorov-Arnold Networks (KAN), and Wavelet KAN. KAN-based models use learnable functions on edges instead of applying fixed  activation functions on nodes/neurons. Traditional KAN and WavKAN mainly differ in the types of functions used. Number of nodes in network layers are mentioned at the bottom. } 
    }
    \label{fig:layer_visualization}
\end{figure*}

\subsubsection{KAN-based Convolutions}
The natural idea of extending KAN concept to convolutions is to replace the scalar product in the convolution operation used in CNNs with a learnable non-linear activation function applied to each element. Given the pre-existence of CNNs, this extension naturally emerges from the foundational concept of KANs. Hence, we avoid associating credit of ConvKANs to any specific contribution. We can define the convolutional kernel implemented for ConvKANs as: 
\begin{equation}
\text{ConvKAN Kernel} = \left[ \begin{array}{cc} \phi_{11} & \phi_{12} \\ \phi_{21} & \phi_{22} \end{array} \right],
\end{equation}
and the computation of a KAN-based convolutional kernel is given by:
\begin{equation}
y_{i,j} = \sum_{m,n} \phi_{m,n}(x_{i+m,j+n}),
\end{equation}
where $\phi_{m,n}$ are learnable functions parameterized as splines. Henceforth, for brevity, we refer to a KAN-based Convolutional layer as KANConv layer. Note the difference between ConvKAN, which denotes a network and KANConv, which is used to identify a  convolutional layer. We abbreviate KANConv as KConv for KConvKANs.

\subsection{Further Details} 
\subsubsection{KANConv Layer}
Consider a KAN layer with input $\mathbf{x} \in \mathbb{R}^{n_{\text{in}}}$ and output $\mathbf{y} \in \mathbb{R}^{n_{\text{out}}}$. The layer applies a matrix of learnable functions $\boldsymbol{\Phi}$:
\begin{equation}
\mathbf{y} = \boldsymbol{\Phi}(\mathbf{x}),
\end{equation}
where $\boldsymbol{\Phi} = \left\{ \phi_{q,p} \right\}$ and $\phi_{q,p}$ are parameterized as B-splines.
On  similar lines,  consider a KANConv layer with input $\mathbf{x} \in \mathbb{R}^{H \times W \times C_{\text{in}}}$ and output $\mathbf{y} \in \mathbb{R}^{H' \times W' \times C_{\text{out}}}$. \textcolor{black}{The layer applies a set of KAN based kernels $\boldsymbol{\Phi}$:} 
\begin{equation}
y_{i,j,c} = \sum_{m,n} \phi_{m,n,c}(x_{i+m,j+n})
\end{equation}
where $\phi_{m,n,c}$ are parameterized as B-splines.

For a $K \times K$ kernel, each element of the kernel has a learnable function $\phi$ with parameter count $G + 2$, where $G$ is the grid size. The total parameter count for a KANConvs layer is $K^2(G + 2)$.

Based on the original proposal~\cite{liu2024kan}, each learnable function $\phi$ in KANs is parameterized as a B-spline:
\begin{equation}
\phi(x) = w_1 \cdot \text{spline}(x) + w_2 \cdot b(x),
\end{equation}
where $b(x)$ is a basis function (e.g., SiLU), and $\text{spline}(x)$ is a linear combination of B-splines. 

KANConv layers provide a flexible and adaptive filtering process, enabling nuanced and controllable transformations of input data, which may lead to learning more complex patterns efficiently. Figure~\ref{fig:layer_visualization} provides a high-level comparison to understand the structural composition of MLP, KAN, and WavKAN.  An architecture  based entirely on KAN Linear layers and KANConv layers is termed  KConvKAN herein. 

\subsection{Wavelet KANs (WavKANs)}

The Wavelet Transform \cite{Gao2011,Gao2011_a} is utilized in signal processing to analyze the frequency content of a signal as it varies over time. The wavelet based extension of KANs, WavKAN \cite{bozorgasl2024wavkan} integrates wavelets into the convolutional framework to enhance feature extraction. The continuous wavelet transform employs a fundamental function called the ``mother wavelet." This function acts as a prototype that can be adjusted in scale and position to align with different segments of the signal. The form of the mother wavelet is crucial because it determines which aspects of the signal are more critical. 

WavKAN employs a mother wavelet $\psi \in L^2(\mathbb{R})$ to analyze the signal $g(t) \in L^2(\mathbb{R})$. A mother wavelet must satisfy the following criteria: 
\begin{enumerate}
    \item \textbf{Zero Mean}:
\begin{equation}
\int_{-\infty}^{\infty} \psi(t) \, dt = 0.
\end{equation}

    \item \textbf{Admissibility Condition}:
\begin{equation}
C_\psi = \int_{0}^{+\infty} \frac{|\hat{\psi}(\omega)|^2}{\omega} \, d\omega < +\infty.
\end{equation} where $\hat{\psi}$ is the Fourier transform of the wavelet $\psi(t)$.

\end{enumerate}

The continuous wavelet transform of a function is represented by wavelet coefficients:
\begin{equation}
C(s, \tau) = \int_{-\infty}^{+\infty} g(t) \frac{1}{\sqrt{s}} \psi\left(\frac{t-\tau}{s}\right) dt,
\end{equation}
where:
\begin{itemize}
  \item $g(t)$ is the signal/function to be approximated by the Wavelet basis.
  \item $\psi(t)$ is the mother wavelet.
  \item $s \in \mathbb{R}^{+}$ is the scale factor.
  \item $\tau \in \mathbb{R}$ is the shift factor.
  \item $C(s, \tau)$ measures the match between the wavelet and the signal at scale $s$ and shift $\tau$.
\end{itemize}

A signal can be reconstructed from its wavelet coefficients using the inverse CWT:
\begin{equation}
g(t) = \frac{1}{C_\psi}  \int_{-\infty}^{+\infty} \int_{0}^{+\infty} C(s, \tau) \frac{1}{\sqrt{s}} \psi\left(\frac{t-\tau}{s}\right) \frac{ds\,d\tau}{s^2}.
\end{equation}
where $C_\psi$ is a constant that ensures the reconstruction's accuracy.

In WavKANConv layers, the wavelet transform is integrated into the convolutional operation. Each wavelet-transformed feature is then processed through the KAN framework, allowing for localized wavelets combined with the adaptive learning capabilities of KANs.

A WavKANConv layer operation with input $\mathbf{x} \in \mathbb{R}^{H \times W \times C_{\text{in}}}$ and output $\mathbf{y} \in \mathbb{R}^{H' \times W' \times C_{\text{out}}}$ can be expressed as:

\begin{equation}
y_{i,j_c} =  \sum_{m,n} \phi_{m,n,c} \left( \int x_{i+m, j+n} \cdot \frac{1}{\sqrt{s_c}} \psi_c\left(\frac{t - \tau_{m,n}}{s_c}\right) dt \right),
\end{equation}
where \( x_{i+m, j+n} \) represents the input data at position \( (i+m, j+n) \), \( \psi_c(t, s_c, \tau_{m,n}) \) denotes the mother wavelet function associated with channel \( c \), \( \frac{1}{\sqrt{s_c}} \) is the normalization factor for the wavelet function, and \( \phi_{m,n,c} \) are learnable functions. 
\begin{figure*}[t]
    \centering
    \includegraphics[width=\textwidth]{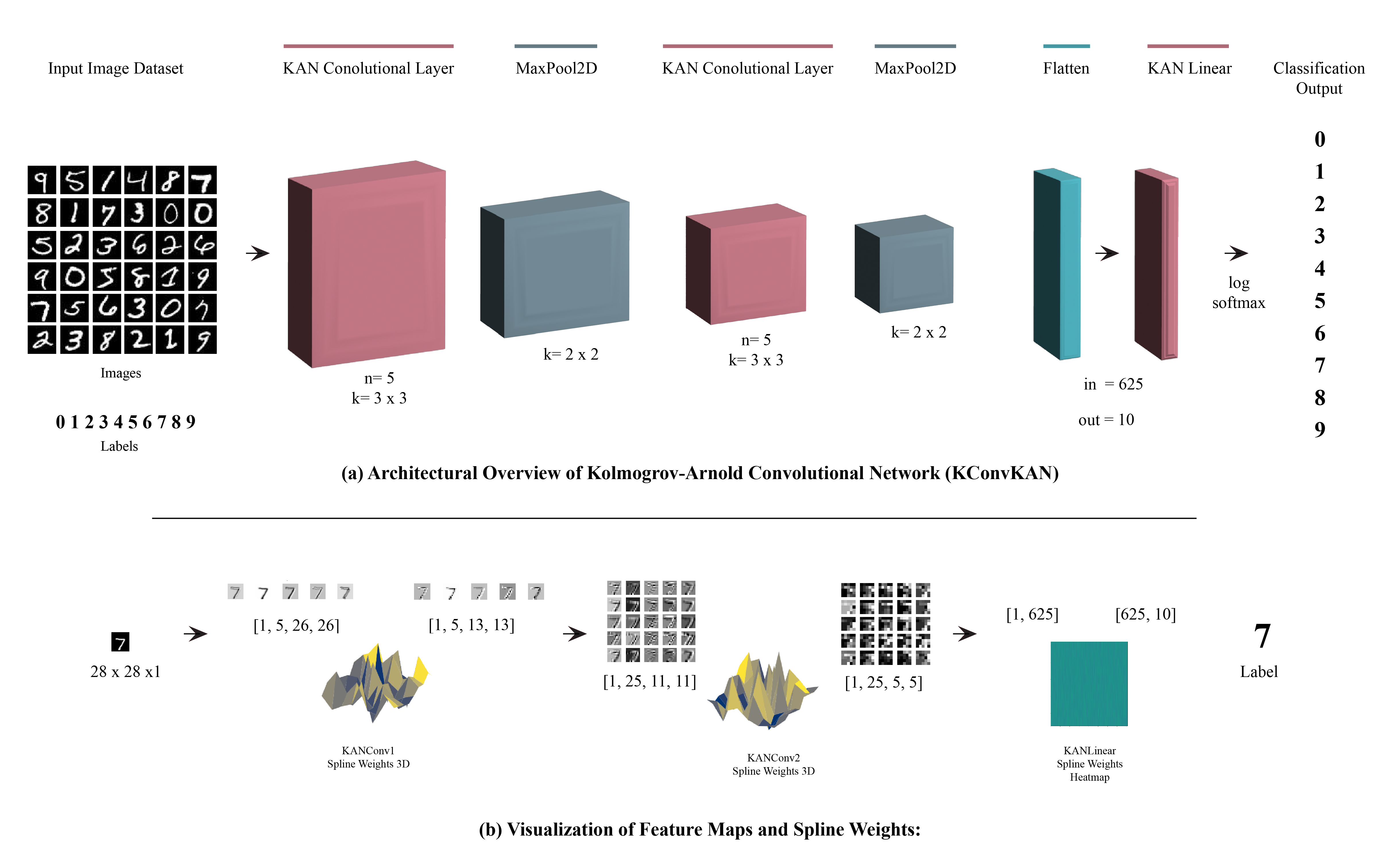}
    \caption{\textbf{(a)} Architectural overview of a KConvKAN used in our experiments to classify on MNIST dataset. \textbf{(b)} Visualization of feature maps and spline weights for the corresponding layer in \textbf{(a)}.}
    \label{fig:kan_visualization}
\end{figure*}
WavKANConvs provide a powerful extension to KANs, incorporating wavelet-based frequency analysis into the adaptive filtering process of convolutional networks. This combination enables the extraction of complex patterns with enhanced localization in both time and frequency domains, potentially leading to superior performance in various visual tasks. 
\subsection{KANs in Segmentation}
\textcolor{black}{Segmentation refers to partitioning an image into meaningful segments at the pixel level, to distinguish different objects or regions. While KANs have shown promise in image recognition tasks, their application to more complex tasks like segmentation has hardly been  explored. The U-Net architecture \cite{ronneberger2015unet} is known for its efficacy in  image segmentation and other pixel-wise prediction tasks. It provides an excellent test bed for analyzing KANs, specifically UKAN \cite{li2024ukan} for segmentation performance.} \textcolor{black}{The motivation behind adapting KANs to segmentation tasks stems from the potential of Kolmogorov-Arnold Networks to learn complex, nonlinear mappings more efficiently than traditional neural networks. By incorporating KAN into the U-Net architecture, the learnable functions on the edges are explored. }

\subsubsection{UKAN-Architecture}
\textcolor{black}{To investigate the applicability of KANs in segmentation, an  adaptation of the U-Net architecture with KAN layers integrated into the convolutional and deconvolutional blocks is adapted from \cite{li2024ukan}, \cite{JaouadTKANUNet}.}

\begin{figure*}[htb]
    \centering
    \includegraphics[width=\textwidth]{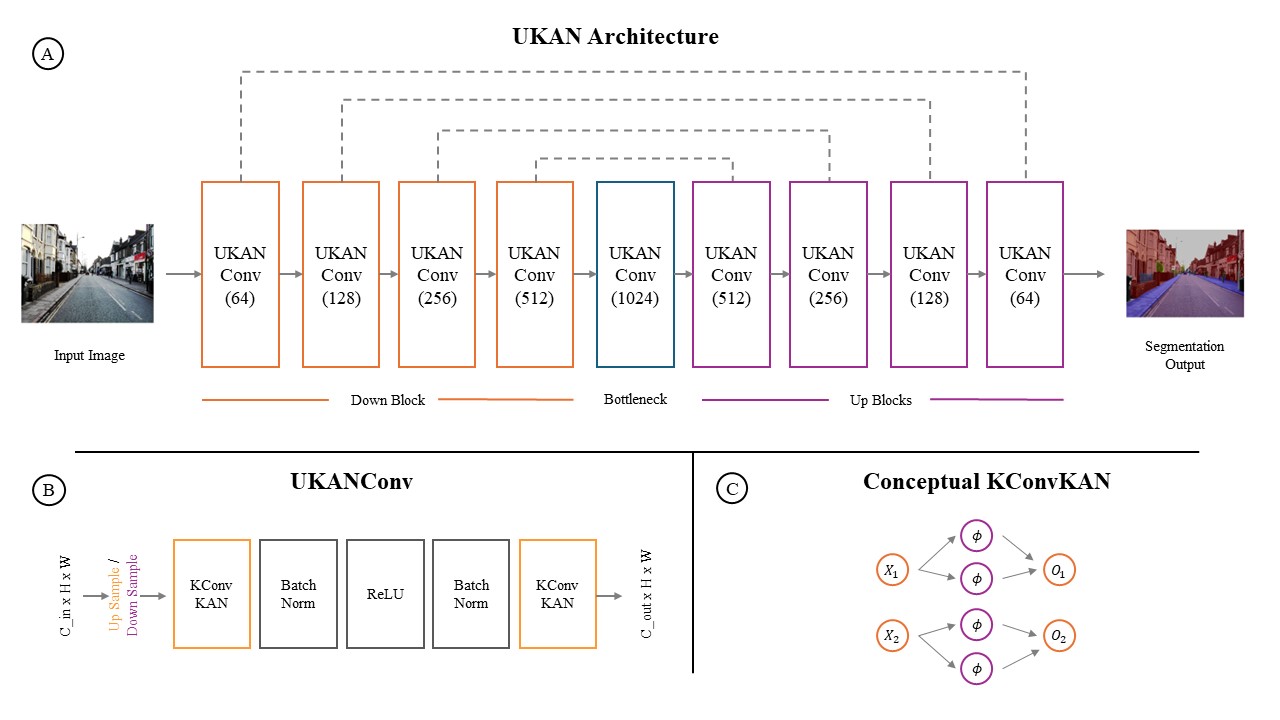}
    \caption{
    \textbf{(A)} Overall UKAN Architecture: The UKAN architecture comprises an encoder (Down Blocks), a bottleneck, and a decoder (Up Blocks) pathway. Each block in the encoder and decoder contains UKAN convolutional layers, which are connected by skip connections (dashed lines) to corresponding layers across the encoder and decoder. The input image is progressively downsampled and then upsampled to produce the segmentation output.
    \textbf{(B)} UKANConv Block: This block, used within the Down and Up blocks, performs two consecutive KConv KAN operations, each followed by Batch Normalization and ReLU activation, which together transform the input feature maps.
    \textbf{(C)} Conceptual KConvKAN: The KConv KAN operation involves applying learnable spline functions (denoted by \( \phi \)) to the input features \( x_1 \) and \( x_2 \), producing output features \( o_1 \) and \( o_2 \). 
    }
    \label{fig:ukan_architecture}
\end{figure*}
The structure of the network is provided as:
\begin{itemize}
    \item \textbf{Encoder:} \textcolor{black}{The encoder in UKAN-UNet follows the typical U-Net design, consisting of convolutional blocks followed by max-pooling layers. However, instead of standard convolutions, we use UKAN layers. These layers replace the conventional linear transformations with learnable, nonlinear spline functions on the edges}
    \item \textbf{Bottleneck:} \textcolor{ black}{The bottleneck, which connects the encoder and decoder, also employs UKAN layers, ensuring that the most abstracted feature maps benefit from the enhanced representational power of KANs.}
    \item \textbf{Decoder:} \textcolor{black}{In the decoder, UKAN layers are used in the upsampling blocks, allowing the network to reconstruct the spatial details with greater accuracy. The skip connections between corresponding encoder and decoder layers ensure that the high-resolution features are retained, further improved by the flexibility of UKAN.}
\end{itemize}

\subsubsection{Mathematical Formulation}
Let \( X \) represent the input image, and \( Y \) the output segmentation map. In a traditional U-Net, the convolutional operations can be represented as:
\[
Y = f_{\text{conv}}(X)
\]
where \( f_{\text{conv}} \) denotes the convolutional layers.

In UKAN-UNet, the convolutional operations are replaced by UKAN layers:
\[
Y = f_{\text{ukan}}(X)
\]
where \( f_{\text{ukan}} \) represents the UKAN layers, defined as:
\[
f_{\text{ukan}}(X) = \sum_{m,n} \phi_{m,n}(X_{i+m,j+n})
\]
Here, \( \phi_{m,n} \) are the learnable functions parameterized as splines, allowing the network to adapt more flexibly to the input data.


\begin{table*}[h]
\centering
\caption{KANLinear Layer Details}
\label{tab:kanlinear}
\begin{tabular}{|c|c|c|c|c|c|c|}
\hline
\textbf{Model Type} & \textbf{In/Out Features} & \textbf{Grid Size} & \textbf{Spline Order} & \textbf{Scale Noise} & \textbf{Activation Function} & \textbf{Grid Range} \\
\hline
KAN Linear & 625/10 & 5 & 3 & 0.1 & SiLU & [-1, 1] \\
\hline
\end{tabular}
\end{table*}

\begin{table*}[h]
\centering
\caption{Convolutional KAN (ConvKAN) Layer Details}
\label{tab:convkan}
\begin{tabular}{|c|c|c|c|c|c|}
\hline
\textbf{Model Type} & \textbf{Grid Size} & \textbf{Spline Order} & \textbf{Scale Noise} & \textbf{Activation Function} & \textbf{Kernel Size/Stride} \\
\hline
KANConv Layer & 5 & 3 & 0.1 & SiLU & (3,3)/(1,1) \\
\hline
\end{tabular}
\end{table*}

\begin{table*}[h]
\centering
\caption{Conceptual-level compositions of networks used in experiments.}
\label{tab:combinedkan}
\begin{tabular}{|c|c|c|c|}
\hline
\textbf{Model Type} & \textbf{Layer Types} & \textbf{Activation Functions} & \textbf{KAN Layers} \\
\hline
ConvKANLinear & Convolutional Layers \& KAN Linear Layer & Multiple & KANLinear \\
KConvLinear & KConv Layers \& Fully Connected Layers & Multiple & KConvs \\

\hline
\multicolumn{4}{c}{} \\[-1ex] 

\hline
\textcolor{black}{KConvKAN} & \textcolor{black}{KConv Layers \& KAN Linear Layers} & \textcolor{black}{SiLU} & \textcolor{black}{KConvs, KANLinear} \\

\hline

\end{tabular}
\end{table*}
\section{Experiments}
\textcolor{black}{This section details the experimental setup and procedures used to evaluate the performance of networks on  MNIST and CIFAR-10 datasets for classification and Camvid dataset for segmentation.} The goal of these experiments is to assess the suitability of KANs for the fundamental computer vision task of recognition, focusing on aspects such as accuracy, model size, training time, and parameter efficiency. \textcolor{black}{The architectural overview and the internal visualizations of a KConvKAN used in our experiments are illustrated in Fig.~\ref{fig:kan_visualization} for a clear understanding. The shown KConvKAN in Fig.~\ref{fig:kan_visualization}(a) depicts the flow of data through the network that consists of two KAN Convolutional (KANConv) layers, each followed by a MaxPool2D layer, then a Flatten layer, and finally a KAN Linear layer that maps the flattened output of the convolutional layers to the final classification output via a log-softmax function. Fig. ~\ref{fig:kan_visualization}(b) provides insights into the feature maps and learned spline weights at different stages of the KConvKAN when processing a sample image}. The feature map  progression is shown after each KANConv and MaxPool2D layer. The 3D spline weights and heat map representation of spline weights in the KAN Linear layer are also visualized. The visualization is meant to demonstrate the transformation of the input through the network layers for understanding. 
\subsection{KAN in Classification}
\subsubsection{Experimental Setup}
The experiments are carried out on a single node of the Spartan High Performance Computing (HPC) system at the University of Melbourne, equipped with 4 NVIDIA A100 GPUs (80GB each), 495GB of RAM, and 32 CPU cores. One GPU was allocated from the gpu-a100 partition for each experiment, ensuring robust computational support for our deep learning models.
In our experiments, we utilized the AdamW optimizer with a learning rate of \(1 \times 10^{-3}\) and varying levels of weight decay to enhance regularization. Training sessions spanned several epochs, managed under an exponential learning rate scheduler with a gamma of 0.8, effectively reducing the learning rate each epoch to refine convergence. The models were evaluated using the cross-entropy loss across classes.
We have replicated the `Convolutional-KANs`\footnote{https://github.com/AntonioTepsich/Convolutional-KANs} \cite{AntonioTepsich2024} and `torch-conv-kan'\footnote{https://github.com/IvanDrokin/torch-conv-kan} \cite{IvanDrokin2024} implementations to evaluate their performance on the MNIST and CIFAR-10 datasets. Our replication efforts note the following. The `Convolutional-KANs' implementation demonstrated adaptation to Convolutional KANs.
 The `torch-conv-kan' implementation provided framework for applying KANs in convolutional layers.

\subsubsection{Datasets}
In this version, the models are evaluated on two datasets widely recognized in the machine learning community for evaluating image recognition models. 
\begin{itemize}
    \item \textbf{MNIST Dataset:} This dataset includes 70,000 handwritten digits, divided into a training set of 60,000 images and a test set of 10,000 images. Each image is grayscale and has a resolution of 28x28 pixels.
    \item \textbf{CIFAR-10 Dataset:} Comprising 60,000 color images across 10 classes, with each class representing a different type of object (e.g., cars, birds), this dataset is split into a training set of 50,000 images and a test set of 10,000 images. Each image is a three-channel RGB color with 32x32 pixels, offering a more challenging task due to its color content and the greater complexity of the images compared to MNIST. 
\end{itemize}

\begin{table*}[h]

\centering
\caption{Overall Performance for Different  Models on MNIST and CIFAR-10.}
\label{tab:performance_metrics_both}
\renewcommand{\arraystretch}{1.5} 
\begin{tabular}{@{}>{\bfseries}l*{8}{c}@{}}
\toprule
\textbf{Model} & \multicolumn{4}{c}{\textbf{MNIST (\%)}} & \multicolumn{4}{c}{\textbf{CIFAR-10 (\%)}} \\
\cmidrule(lr){2-5} \cmidrule(lr){6-9}
 & \textbf{Accuracy} & \textbf{Precision} & \textbf{Recall} & \textbf{F1 Score} & \textbf{Accuracy} & \textbf{Precision} & \textbf{Recall} & \textbf{F1 Score} \\
\midrule
SimpleMLP & 92.4 & 92.4 & 92.3 & 92.3 & 39.1 & 39.5 & 39.4 & 39.1 \\
ConvNet (Small) & 98.4 & 98.4 & 92.3 & 92.3 & 56.2 & 55.9 & 56.2 & 55.7 \\
ConvNet (Medium) & 99.1 & 99.1 & 99.1 & 99.1 & 64.2 & 64.4 & 64.2 & 64.2 \\
ConvNet (Large) & 99.4 & 99.4 & 99.4 & 99.4 & 71.0 & 71.3 & 71.0 & 70.8 \\
ConvKANLinear & 98.5 & 98.5 & 98.5 & 98.5 & 61.6 & 61.5 & 61.6 & 61.4 \\
KConvLinear & 98.3 & 98.3 & 98.3 & 98.3 & 59.3 & 59.3 & 59.3 & 59.2 \\
\midrule

{KConvKAN (2 Layers)} & 98.8 & 98.8 & 98.8 & 98.8 & 62.6 & 62.4 & 62.6 & 62.3 \\
\midrule
{KConvKAN (8 Layers)} & 99.6 & 99.6 & 99.5 & 99.6 & 78.8 & 78.6 & 78.5 & 78.6 \\
\midrule
WavKan (2 Layers)& 98.8  & 98.8 & 98.7 & 98.8 & 64.4 & 64.4 & 66.3 & 66.1 \\
\midrule
WavKan (8 Layers) & 99.6  & 99.6 & 99.5 & 99.6 & 79.7 & 79.4 & 79.5 & 79.4 \\
\bottomrule
\end{tabular}

\end{table*}

\subsubsection{Model Configurations}
In our experiments, we implemented several configurations of KANs,  ConvsKANs, WavKAN, MLPs and traditional ConvNets to assess  performance. 
\begin{itemize}
    \item \textbf{Baseline Models:} We used traditional MLPs and CNNs as baselines to establish a reference for evaluating the KANs and ConvKANs.
    \item \textbf{KANs with Traditional Layers:} We explored combinations of KAN layers with traditional convolutional and MLP layers to assess performance across different network architectures. Table~\ref{tab:kanlinear} provides the configurations for our KANLinear layers, which are designed to test the flexibility and efficiency of KANs in processing linear layers with standard parameters. Table~\ref{tab:combinedkan} provides compositional details of how we have integrated KAN layers with traditional neural network layers to test the hybrid models. 
    \item \textbf{KANConv Implementations:} Baseline configurations of KANConv layers were tested to examine their effects on convolutional processing, particularly for feature extraction and classification accuracy. Table \ref{tab:combinedkan} outlines high-level composition of our KConvKAN models. 
\end{itemize}

In this study, we evaluate numerous neural network architectures incorporating KAN principles. These configurations are designed to assess their utility without any bias toward specific model enhancements.

\subsubsection{Results}
This section presents the performance evaluation of previously introduced network models trained on the MNIST and CIFAR-10 datasets. The evaluation focuses on key metrics such as accuracy, precision, recall, and F1 score. In addition, comparisons of parameters and evaluation time are presented. The comparative analysis of model performance on the MNIST and CIFAR-10 datasets reveals significant variations across different architectures, as detailed in Table \ref{tab:performance_metrics_both}. Models demonstrate high precision, recall, and F1 scores on MNIST.
In contrast, performances on CIFAR-10 are considerably lower, underscoring the complexity of the dataset. 
Table \ref{tab:parameters} present the evaluation in terms of accuracy values and the parameters of each individual mode. The results indicate that models with increased parameter counts generally show higher accuracy on both datasets. 
This pattern suggests that the complexity of the models may play a significant role in handling the challenges presented by the respective datasets.

The KAN architectures, such as KConvKAN and ConvKANLinear, evaluated on MNIST and CIFAR-10, as highlighted in Table \ref{tab:parameters}, showcase varied effectiveness. For MNIST, KConvKAN (2 Layers) achieves an accuracy of 98.8\%, which is competitive  but not the highest. However, KConvKAN with 8 layers achieves 99.6\% results along WavKAN (8 layers). On CIFAR-10, the performance  increases with more specialized KAN models, indicating a nuanced benefit in handling more complex image tasks. While these KAN-enhanced models do not always lead in performance metrics, they maintain consistently high scores across the board. 

\begin{table}[t]
\begin{threeparttable}
\centering
\caption{Model Performance and Parameters}
\label{tab:parameters}
\begin{tabular}{@{}llccc@{}}
\toprule
Dataset & Model & Accuracy (\%) & Parameters & Epochs \\ 
\cmidrule(r){1-5}
\multirow{7}{*}{MNIST} & SimpleMLP & 92.4 & 7,850 & \multirow{7}{*}{10} \\
& ConvNet (Small) & 98.4 & 2,740 & \\
& ConvNet (Medium) & 99.1 & 157,030 & \\
& ConvNet (Large) & 99.4 & 887,530 & \\
& ConvKANLinear & 98.5 & 163,726 & \\
& KConvLinear & 98.3 & 37,030 & \\
& {KConvKAN} & 98.8 & 94,650 & \\ \midrule
\multirow{3}{*}{MNIST} & {KConvKAN-8} & 99.6 & 40,694,018 & \multirow{3}{*}{150} \\
\cmidrule(r){2-4}
& WavKAN-2 & 98.8 & 951,370 & \\
\cmidrule(r){2-4}
& WavKAN-8 & 99.6 & 10731562 & \\ \midrule
\multirow{7}{*}{CIFAR-10} & SimpleMLP & 39.5 & 30.730 & \multirow{7}{*}{10} \\
& ConvNet (Small) & 56.2 & 3,580 & \\
& ConvNet (Medium) & 64.2 & 224,495 & \\
& ConNet (Large) & 71.0 & 1,134,890 & \\
& ConvKANLinear & 61.6 & 694,926 & \\
& KConvLinear & 59.3 & 48,370 & \\
& {KConvKAN} & 62.6 & 405,900 & \\ \midrule
\multirow{3}{*}{CIFAR-10} & {KConvKAN-8} & 78.8 & 40,695,584 & \multirow{3}{*}{150} \\
\cmidrule(r){2-4}
& WavKAN-2 & 64.4 & 952,650 & \\
\cmidrule(r){2-4}
& WavKAN-8 & 79.7 & 10,732,202 & \\ 
\bottomrule
\end{tabular}
\begin{tablenotes}
\item The variation in epochs across models is intended to investigate the effect of more extensive training durations on performance of KAN based models.
\end{tablenotes}
\end{threeparttable}
\end{table}
\begin{table}[h]
\centering
\caption{Evaluation Time in Comparison with Accuracies}
\label{tab:model_time}
\begin{tabular}{lclcl}
\toprule
Model & Dataset & Accuracy & Train Time & Eval. Time \\
\midrule
KConvKAN-8 & \multirow{3}{*}{MNIST} & 99.6\% & 1hr-55min & 4.75 sec\\
WavKAN-2 &  & 98.8\% & 3hr-33min & 5.72 sec\\
WavKAN-8 &  & 99.6\% & 8hr-40min & 13.76 sec\\
\midrule
KConvKAN-8 & \multirow{3}{*}{CIFAR-10} & 78.8\% & 2hr-05min& 5.20 sec\\
WavKAN-2 &  & 64.4\% & 3hr-45min& 7.81 sec\\
WavKAN-8 &  & 79.7\% & 8hr-55min& 12.92 sec\\
\bottomrule
\end{tabular}
\begin{tablenotes}
\item The batch size used for WavKAN and WavKAN8 is 64, and for KConvKAN-8 it is 128, consistent across both datasets.
\end{tablenotes}
\end{table}

\begin{figure*}[ht]
    \centering
    \begin{minipage}{0.5\textwidth}
        \centering
        \includegraphics[width=0.95\linewidth]{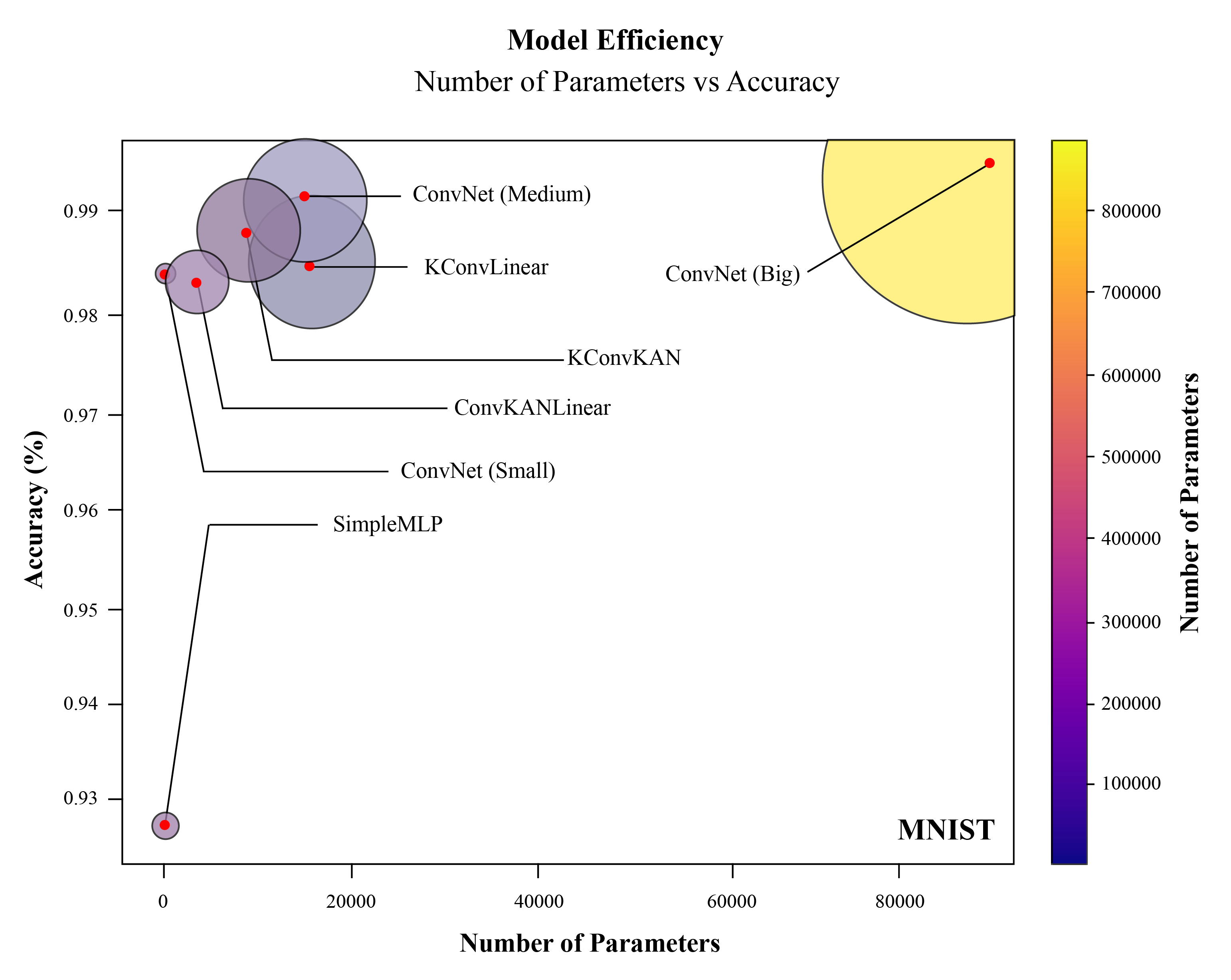}
    \end{minipage}%
    \begin{minipage}{0.5\textwidth}
        \centering
        \includegraphics[width=0.95\linewidth]{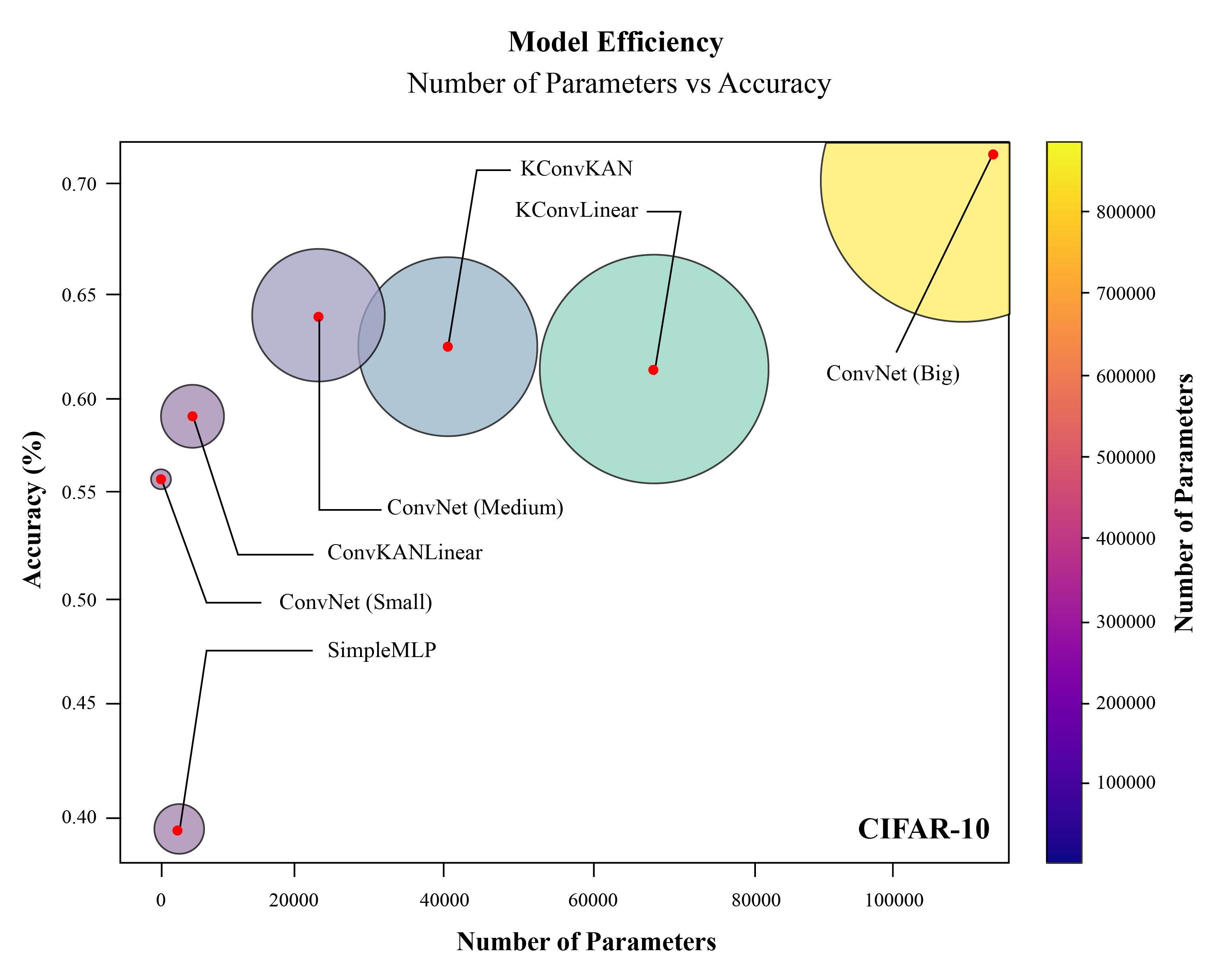}
    \end{minipage}
    \caption{Comparative performance visualization of different models on  MNIST and CIFAR-10 datasets. The sizes and colors of the circles represent the scale of model parameters and performance metrics respectively, showcasing a range of outcomes from simple to complex models.}
    \label{fig:both_performance}
\end{figure*}

In Fig.~\ref{fig:both_performance}, the relationship between the number of parameters in each model and the achieved accuracy on both datasets is explored. The graphical representation uses circle sizes to denote parameter count and color intensity to reflect accuracy, offering a visual correlation between model complexity and performance. The KAN architectures, positioned between the simplest and most complex models, illustrate a balanced approach to leveraging model design for effective learning. This balance might hint at potential efficiencies in parameter usage while achieving competitive accuracies, especially noticeable in the CIFAR-10 dataset where higher model complexities generally correlate with better performance.



\subsection{KAN in Segmentation}

\subsubsection{Experimental Setup}
\textcolor{black}{The computation experimental setup is same as outlined in the classification section. The UKAN-UNet was implemented using the PyTorch framework. We initialized the UKAN layers with a grid size of 5 and a spline order of 3. The training was performed using an Adam optimizer with an initial learning rate of 0.001, and the learning rate was decayed exponentially with a gamma of 0.8.} The model was trained for \textcolor{black}{1000} epochs with a batch size of 16 on an NVIDIA A100 GPU.  
\subsubsection{Dataset}
\textbf{CamVid:} \textcolor{black}{The CamVid dataset, or Cambridge-driving Labeled Video Database, is a widely used benchmark in the field of semantic segmentation, particularly for autonomous driving applications \cite{brostow2009semantic}. This dataset comprises over 700 images extracted from five high-resolution video sequences captured at 30 frames per second using a camera mounted on a vehicle's dashboard. These sequences were then downsampled to provide labeled still frames at 1 frame per second, resulting in a total of 701 images. Each image is annotated with pixel-level labels corresponding to 32 distinct semantic classes, including categories such as sky, building, road, tree, and pedestrian, among others \cite{sturgess2009combining}. For practical purposes, these classes are often condensed into 11 primary categories to facilitate research and model training. The CamVid dataset is particularly valuable for its high-quality annotations and the diversity of urban driving scenes it represents, making it an essential resource for developing and evaluating semantic segmentation algorithms in dynamic environments \cite{badrinarayanan2017segnet}.}

\subsubsection{Results}
\textcolor{black}{The performance of UKAN-UNet was evaluated using standard segmentation metrics, including pixel accuracy, mean Intersection over Union (mIoU), and Dice coefficient.} 
\begin{table*}[t]
    \centering
    \caption{Test Loss Values for UKAN and UNet at 500 and 1000 Epochs On CamVid Dataset}
    \begin{tabular}{lcccc}
        \toprule
        \textbf{Metric} & \textbf{500 Epochs - UKAN} & \textbf{500 Epochs - UNet} & \textbf{1000 Epochs - UKAN} & \textbf{1000 Epochs - UNet} \\
        \midrule
        Test Loss (Memory)   & 0.9550 & 1.0487 & 0.8443 & 0.8991 \\
        \midrule
        Test Loss (Saved)   & 0.9800 & 1.0401 &  0.8316 & 0.8872 \\
        \midrule
        Test Loss (Final Evaluation)    & 0.9758 & 1.0474 & 0.8296 & 0.8637 \\
        \bottomrule
    \end{tabular}
    \label{tab:summarized_loss}
\end{table*}

\begin{table}[t]
    \centering
    \caption{Segmentation Results on the CamVid Dataset}
    \begin{tabular}{lcc}
        \toprule
        \textbf{Metric} & \textbf{UKAN-UNet} & \textbf{U-Net} \\
        \midrule
        Pixel Accuracy & 67.3\% & 64.5\% \\
        mIoU          & 57.8\% & 54.2\% \\
        Dice Coefficient & 64.7\% & 61.4\% \\
        \bottomrule
    \end{tabular}
    \label{tab:segmentation_results}
\end{table}

\begin{itemize}
    \item \textbf{Accuracy:} UKAN-UNet achieved a pixel accuracy of \textcolor{black}{67.3\%} on the CamVid dataset, compared to \textcolor{black}{64.5\%} for the baseline U-Net.
    \item \textbf{mIoU:} The mean IoU improved from \textcolor{black}{54.2\%} with U-Net to \textcolor{black}{57.8\%} with UKAN-UNet, indicating better segmentation across all classes.
    \item \textbf{Dice Coefficient:} UKAN-UNet also showed an improvement in the Dice coefficient, overall.
\end{itemize}

\subsection{Claims Validation}
Kolmogorov-Arnold networks~\cite{liu2024kan}  are proposed as alternates to MLPs, and by extension, to other architectures like CNNs. Hence, their claims regarding accuracy, interpretability, scalability, and  computational efficiency require validation for computer vision applications. In this section, we briefly discuss these aspect while comparing KANs and MLPs.

\subsubsection{Claim-1: Accuracy}

The original work on KANs~\cite{liu2024kan}  identifies that smaller KAN models can match or exceed the accuracy of larger MLPs in function fitting tasks. We can analyze this claim through our experiments on  MNIST, CIFAR10 and CamVid datasets. Our preliminary results show that KANs show competitive results with ConvNets both in terms of classification and segmentation.  


\subsubsection{Claim-2: Computational Efficiency}
Despite initial concerns about the potential computational overhead of implementing spline functions at scale, our implementations of KANs on standard hardware demonstrated a manageable increase in computational demands compared to MLPs. This is consistent with the original claim~\cite{liu2024kan} that KANs can achieve  comparable performance with potentially lower parameter counts.

\subsubsection{Claim-3: Scalability}
According to the KAN paper, these networks exhibit faster neural scaling laws than MLPs. However, when scaling up the networks to accommodate the high-dimensional data from CIFAR10, KANs required significant tuning of spline parameters to maintain performance, which adds to the training complexity. The adaptation of KANs in UNet architecture provides preliminary support to the scalability claim, suggesting that practical scalability is achievable and contingent on the ability to manage increasing model complexity efficiently. 

\section{Conclusion}
This work critically evaluates the efficacy of Kolmogorov-Arnold networks in the basic computer vision task of image recognition, contrasting their performance with conventional networks across MNIST and CIFAR-10 datasets for classification and CamVid dataset for segmentation. This work is a continuous effort that will be updated as more data and results are available. Our current findings reveal that while KANs do offer variation in the traditional neural network architectures, their superior accuracy and scalability are clearly evident for classification and segmentation datasets. The distinctive architecture of KANs, which integrates learnable spline functions directly onto the edges, provides an innovative approach to neural network design and opens new avenues for enhancing model interpretability and efficiency. The practical application of KANs in complex visual tasks require further optimization and exploration to fully harness their theoretical advantages. The research community is already engaged in leveraging KANs for vision or related tasks. Future work should focus on improving KAN-based architecture for scalability and generalization in diverse application scenarios within computer vision. \textcolor{black}{Overall, the comparative analysis indicates that WavKAN's utilization of wavelet functions not only preserves the strengths of conventional KANs on simpler datasets like MNIST but also extends their applicability and performance on more demanding datasets.} \textcolor{black}{It does support the argument that investing in KAN-based vision models is worthwhile.} With the large interest of the vision and non-vision community in KANs, we can expect to see interesting developments in KANs in terms of computer vision in the near future.
\bibliographystyle{IEEEtran}  
\bibliography{ref}

%

\end{document}